\documentclass[]{bytedance_seed}
\usepackage{makecell}
\usepackage{microtype}
\usepackage{hyperref}
\usepackage{url}
\usepackage{booktabs}
\usepackage{graphicx} 
\usepackage{amsmath}
\usepackage{float}
\usepackage{enumitem}
\usepackage{tabularx}
\usepackage{listings}
\usepackage{array}
\usepackage{subcaption}
\usepackage{fvextra}
\usepackage{booktabs}
\usepackage{siunitx}
\usepackage{multicol}

\usepackage[T1]{fontenc}
\usepackage{lmodern}
\usepackage{pdflscape} 

\usepackage{enumitem}

\usepackage[most]{tcolorbox}
\usepackage{xcolor}
\usepackage{colortbl}
\usepackage{multirow}
\usepackage{bigdelim}  
\makeatletter
\AtBeginDocument{\@ifundefined{CT@arc@}{\let\CT@arc@\relax}{}}
\makeatother

\newtcolorbox{casebox}[2][]{
  enhanced,
  colback=white,
  colframe=black,
  boxrule=1pt,
  arc=3mm,
  left=6pt,right=6pt,top=6pt,bottom=6pt,
  title={#2},
  attach title to upper,   
  #1
}

\newtcblisting{diffblock}{
  listing engine=listings,
  colback=green!5,
  colframe=green!40!black,
  listing only,
  left=2mm,
  right=2mm,
  top=0.5mm,
  bottom=0.5mm,
  boxrule=0.4pt,
  sharp corners,
  enhanced,
  listing options={
    basicstyle=\ttfamily\tiny,   
    columns=fullflexible,
    keepspaces=true
  }
}

\newtcblisting{codebox}{
  listing engine=listings,
  colback=gray!5,
  colframe=gray!75!black,
  listing only,
  left=2mm,
  right=2mm,
  top=0.5mm,
  bottom=0.5mm,
  boxrule=0.5pt,
  arc=2mm,
  enhanced,
  listing options={
    basicstyle=\ttfamily\small,
    columns=fullflexible,
    keepspaces=true
  }
}

\newcounter{case}[section]

\usepackage{lineno}

\definecolor{darkblue}{rgb}{0, 0, 0.5}
\hypersetup{colorlinks=true, citecolor=darkblue, linkcolor=darkblue, urlcolor=darkblue}

\makeatletter
\renewcommand{\sfdefault}{lmss} 

\DeclareRobustCommand{\sffamily}{\fontfamily{\sfdefault}\selectfont}
\makeatother

\title{Agentic Discovery of Exchange--Correlation Density Functionals}

\author[1,2,*]{Titouan Duston}
\author[1]{Jiashu Liang}
\author[1]{Yuanheng Wang}
\author[1]{Weihao Gao}
\author[1]{Xuelan Wen}
\author[1]{Nan Sheng}
\author[1]{Weiluo Ren}
\author[1,\dagger]{Yang Sun}
\author[1,\dagger]{Yixiao Chen}

\affiliation[1]{ByteDance Seed}
\affiliation[2]{Princeton University}

\contribution[*]{Work done at ByteDance Seed}
\contribution[\dagger]{Corresponding authors}

\abstract{
The development of accurate exchange--correlation (XC) functionals remains a longstanding challenge in density functional theory (DFT). The vast majority of XC functionals have been hand designed by human researchers combining physical insight, exact constraints, and empirical fitting. Recent advances in large language models enable a systematic, automated alternative to this human-driven design loop. This report presents an agentic search system in which an LLM proposes structured functional-form changes guided by evolutionary history. The system attempts to improve functional performance through an iterative plan--execute--summarize loop, where improvements are measurable by optimizing functional parameters against a standard thermochemistry dataset, then evaluating performance on a held-out subset. The strongest discovered functional, \texttt{SAFS26-a} (Seed Agentic Functional Search 2026), improves upon the gold-standard $\omega$B97M-V baseline by ${\sim}9\%$. These results also surface a cautionary lesson for AI-assisted science: models powerful enough to discover genuine improvements are equally capable of exploiting unphysical shortcuts to game the benchmark; domain expertise translated into explicitly enforced constraints remains essential to keeping results scientifically grounded.
}

\correspondence{Yang Sun at \email{sunyang.46135@bytedance.com}, Yixiao Chen at \email{yixiao.chen@bytedance.com}}

\begin{document}
\maketitle

\section{Introduction}

The accuracy of Kohn-Sham DFT is determined primarily by the quality of the exchange-correlation functional \(E_{\mathrm{xc}}[\rho]\). Over time, this challenge has produced a dense hierarchy of approximations, from local-density approximations, generalized gradient approximations, and meta-GGAs to hybrids, and double hybrids, each trading off accuracy, cost, and generality. More than 200 XC functionals have been proposed to date, yet there is still no systematic recipe for generating functionals that are reliably accurate across diverse chemical systems~\cite{mardirossian2017thirty}.

This lack of a constructive design procedure makes XC functional development a natural target for automated search. Recent LLM-driven evolutionary systems have been most successful on algorithmic and engineering tasks, where progress is easily measured and successive iterations can realize concrete improvements~\cite{romeraparedes2024funsearch,novikov2025alphaevolve,yang2023opro,wan2025loongflowdirectedevolutionarysearch}. DFT functional discovery, however, poses a fundamentally different challenge. In algorithmic domains, the starting solution is typically far from optimal, and each evolutionary iteration can identify an objective actionable improvement (a better heuristic, a tighter bound, a more efficient subroutine) so that exploitation-heavy parent selection reliably drives progress. In functional discovery, by contrast, many apparent improvements are entangled with caveats that scientists have meticulously determined over decades, including conservation laws, numerical stability, interpretability, and transferability~\cite{kaplan2023predictive}. Moreover, the XC functional baseline is already a highly optimized human design, leaving few local parameter modifications that do not violate physical constraints or regress broader performance. Exploitation-heavy sampling therefore stalls, and repeated micro-perturbation of a single elite tends to overfit the benchmarks rather than improve genuine generalization. Incremental refinement of a near-optimal form is thus not merely ineffective but actively harmful (see Appendix \ref{sec:ablation-creativity}). Genuine progress instead requires first \textit{exploring} structurally novel regions of functional-form space (new descriptor combinations, rational enhancement-factor topologies, and spin-dependent architectures), and only then \textit{refining} within a newly discovered promising region. This exploration-then-refinement dynamic accounts for most of the significant score gains observed in our search (see Appendix \ref{sec:physical-constrained} and \ref{sec:grid-constrained}).

Prior searches over symbolic functional forms can produce competitive functionals~\cite{ma2022evolving}, but semi-random mutations limit the search to a predetermined set of modifications and only weakly exploit accumulated knowledge of what has succeeded or failed. Instead, could a symbolic search be performed with an entirely LLM based agent loop, one that can inspect evolutionary history, diagnose plateaus, and propose qualitatively different forms? More importantly, could these agents actually improve upon the most accurate human-designed functionals? This report focuses on that scientific question. Our goal is to design an agentic search capable of improving upon the SOTA exchange-correlation functional forms, analyze the highest-performing discovered functionals, and extract the design principles and limitations that emerged from the search trajectories.

\subsection{Agentic Discovery Framework}

\begin{figure}[H]
    \centering
    \includegraphics[width=0.60\linewidth]{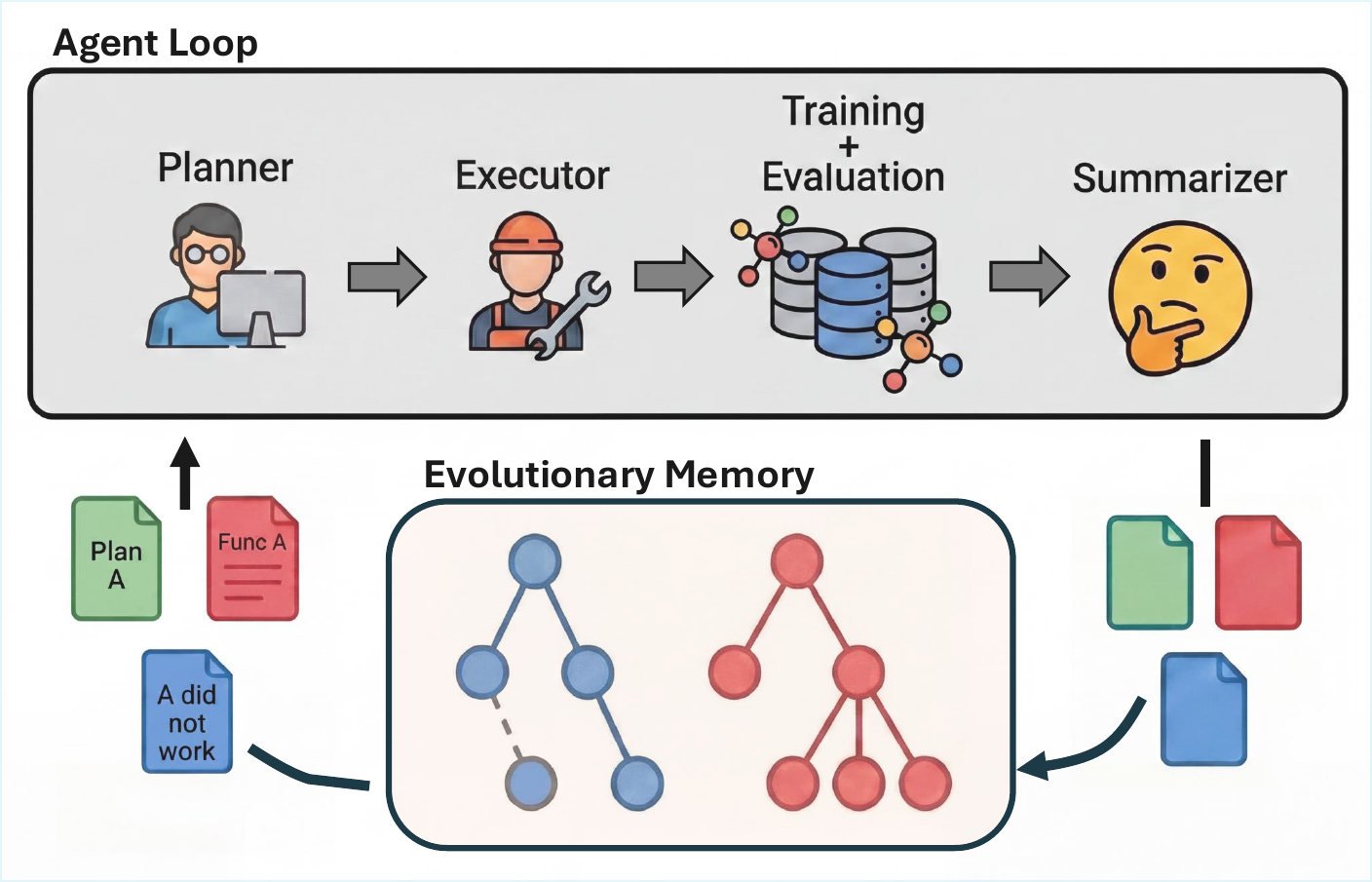}
    \caption{Sketch of the agentic search framework: a Plan--Execute--Summarize loop over a multi-island evolutionary population with structured memory management.}
    \label{fig:framework}
\end{figure}

\textbf{Plan–Execute–Summarize Loop.} The agentic search is built on LoongFlow~\cite{wan2025loongflowdirectedevolutionarysearch}, an evolutionary framework that replaces random mutations with a structured cognitive loop. Each evolutionary iteration passes through three LLM stages: a \textit{Planner} that produces a natural-language blueprint for the next functional-form modification, an \textit{Executor} that translates this blueprint into executable JAX code, and a \textit{Summarizer} that compares intent against outcome and writes a structured diagnosis into the evolutionary memory. The exact task definition and full component descriptions are given in Appendices~\ref{sec:task-definition} and~\ref{sec:evo-framework}.

\textbf{Population Structure.}
\label{sec:pop-structure-intro}
The search maintains a population of candidate functionals across n independently evolving islands, each with a capped archive of top solutions preserving structurally diverse stepping stones~\cite{mouret2015mapelites}. Parent selection uses an exploration-heavy 80/20 split (80\% uniform-random, 20\% from the elite archive), and periodic migration transfers top solutions between islands, enabling cross-lineage fusion of independently matured ideas. This exploration-heavy regime prioritizes structural diversity over exploitation of the current best, an explicit choice to match the nature of the 
problem (Section~\ref{sec:model_dynamics}). Full details are given in Appendix~\ref{sec:pop-structure}.

\textbf{Evolutionary Memory.}
\label{sec:evo-memory-intro}
Over 400+ iterations the accumulated search record grows far beyond any single LLM context window. The framework addresses this through an evolutionary memory \(\mathcal{M}_t\). In addition to the full parent state, search context is delivered to the planner via two channels: per-lineage retrieval tools that surface ancestor plans, solutions, and summarizer diagnoses, and a pre-synthesized global summary of exhausted dead-end strategies compiled across all islands. This ensures agents avoid revisiting failed strategies without requiring the full history to fit in a single prompt. Full details are given in Appendix~\ref{sec:evo-memory}.

\subsection{Functional Search Space And Evaluation}
\textbf{Functional form and search target.} 
The evolving object is the semi-local part of a range-separated hybrid meta-GGA density functional~\cite{ma2022evolving,mardirossian2017thirty}. In a range-separated hybrid meta-GGA, exchange correlation is built from the local spin-densities \(\rho_\sigma\), their gradients \(|\nabla\rho_\sigma|\), and the kinetic-energy densities \(\tau_\sigma\). The total exchange energy combines short-range semi-local exchange, with a fraction \(c_x\) of short-range exact (Hartree-Fock) exchange and full long-range exact exchange. Correlation is likewise decomposed into same-spin and opposite-spin semi-local channels plus a VV10 nonlocal dispersion term~\cite{vydrov2010vv10}:
\begin{align}
    E_{x} &= E_{x,\mathrm{sr}}^{\mathrm{mGGA}}
           + c_x\,E_{x,\mathrm{sr}}^{\mathrm{exact}}
           + E_{x,\mathrm{lr}}^{\mathrm{exact}}, \\
    E_{c} &= E_{c,\mathrm{ss}}^{\mathrm{mGGA}}
           + E_{c,\mathrm{os}}^{\mathrm{mGGA}}
           + E_{c,\mathrm{nl}}^{\mathrm{VV10}}.
\end{align}
Each semi-local channel is defined by multiplying the appropriate local density approximation (LDA) reference energy density by a dimensionless enhancement factor \(g(w,u)\); the form of these enhancement factors is the object of the evolutionary search. The range-separation parameter \(\omega=0.3\), the short-range exact-exchange fraction \(c_x=0.15\), and the VV10 parameters are inherited from \(\omega\)B97M-V and held fixed throughout; only the three enhancement factors \(g_x\), \(g_{c,\mathrm{ss}}\), and \(g_{c,\mathrm{os}}\) are evolved.

The raw densities \(\rho_\sigma\) enter through the LDA reference energy densities that the enhancement factors multiply. Since \(|\nabla\rho|\) and \(\tau\) are unbounded and not dimensionless, the gradients and kinetic-energy densities enter the enhancement factors as dimensionless descriptors,
\begin{equation}
    s = |\nabla \rho| / \rho^{4/3},
    \qquad
    t = \tau / \rho^{5/3},
\end{equation}
where \(\tau = \tfrac{1}{2}\sum_i|\nabla\varphi_i|^2\) is the positive-definite kinetic energy density. The bounded variables are
\begin{equation}
    w = \frac{k_\sigma - t}{k_\sigma + t},
    \qquad
    u = \frac{\gamma s^2}{1+\gamma s^2},
\end{equation}
with \(w\in[-1,1]\) and \(u\in[0,1]\), where \(k_\sigma = \tfrac{3}{10}(6\pi^2)^{2/3}\) is the UEG kinetic-energy prefactor and \(\gamma\) is a fixed gradient parameter inherited from \(\omega\)B97M-V.

\textbf{Baseline and evaluator.} 
The initial baseline is \(\omega\)B97M-V, the gold standard range-separated hybrid meta-GGA density functional~\cite{mardirossian2016wB97MV,mardirossian2017thirty}. In the baseline form, exchange is represented by a compact polynomial in \(w\) and \(u\), and correlation is represented by low-order polynomials in spin-resolved or averaged descriptors. Candidate descendants are implemented in a JAX-based codebase that evaluates functional forms on electron densities held fixed at the \(\omega\)B97M-V self-consistent values, following a non-self-consistent evaluation protocol~\cite{ma2022evolving}. The empirical parameters of each candidate are fitted by minimizing the training error using a quasi-Newton optimization algorithm (L-BFGS), with gradients computed via auto-differentiation through JAX. Loss is measured through the weighted root mean squared deviation (WRMSD) on the MGCDB84 dataset. To avoid leaking the final test set in the evolution, evaluation uses the validation subset WRMSD to calculate the score. Following the procedure established for \(\omega\)B97M-V, final performance is measured by the WRMSD computed over the combined validation and test subsets of MGCDB84, excluding the RG10 dataset. RG10 is excluded because it does not belong to any of the eight thermochemical data types into which the remaining 82 datasets are classified; in total 3547 points make up this final set~\cite{mardirossian2016wB97MV}. MGCDB84 training-split was minimally changed with the transfer of two molecules from the test set to the validation set (C20C24 subset). This was done because this subset is known to be significantly out-of-distribution; thus, its inclusion added a critical evolution signal.

Because each candidate is evaluated on a density that is not self-consistent for it, the reported WRMSD values may shift once the density is allowed to relax. Ma~et~al.~\cite{ma2022evolving} found that self-consistent GAS22 errors were close to their non-self-consistent counterparts, but analogous SCF validation remains a necessary follow-up.

\textbf{Scoring.}
\label{sec:scoring}
Let \(s\) denote a candidate functional form together with its optimized empirical parameters \(\Theta\). Parameters are fitted on the MGCDB84 training split, and every candidate is scored by the validation weighted root-mean-square deviation
\begin{equation}
J[s(\Theta)]=\mathrm{WRMSD}_{\mathrm{val}}[s(\Theta)], \qquad R[s(\Theta)]=\frac{J_{\mathrm{target}}}{J[s(\Theta)]}.
\end{equation}
Here \(J_{\mathrm{target}} = 3.45\) kcal is an arbitrarily chosen target approximately 15\% below the \(\omega\)B97M-V validation WRMSD (4.02 kcal). To discourage unphysical solutions, the raw score is penalized for violations of the enforced physical constraints (Section~\ref{sec:constraints}):
\begin{equation}
R_{\mathrm{evlv}}[s(\Theta)] = R[s(\Theta)] \times 0.9^{\,n},
\end{equation}
where \(n\) is the number of constraints violated by the candidate. The search therefore favors functional forms whose parameters can be optimized successfully, that generalize to the held-out validation subset, and that satisfy the enforced physical constraints. The test split is reserved for final reporting only.

\subsection{Verification and Robustness}
\label{sec:constraints}

Beyond numerical robustness, the discovered functionals are screened against three physical constraints + one technical constraint that any well-behaved meta-GGA exchange-correlation functional should satisfy~\cite{kaplan2023predictive}. Results collected without penalizing violations of these constraints reveal enforcement is not merely precautionary but forms  essential guardrails (see Appendix \ref{sec:unconstrained}).

\textbf{1. Spin symmetry.}
The XC energy density must be invariant under exchange of spin channels: \(\varepsilon_{\mathrm{xc}}(\mathbf{r};\rho_\alpha,\rho_\beta) = \varepsilon_{\mathrm{xc}}(\mathbf{r};\rho_\beta,\rho_\alpha)\) at every point in space. This was tested on an O\(_2\) triplet molecular grid (where \(\rho_\alpha \neq \rho_\beta\)) by swapping the spin channels and comparing the resulting energy densities pointwise. The maximum pointwise deviation \(\max_{\mathbf{r}}|\varepsilon_{\mathrm{xc}}^{\alpha\beta} - \varepsilon_{\mathrm{xc}}^{\beta\alpha}|\) must fall below \(10^{-5}\)~Ha.

\textbf{2. Uniform electron gas (UEG) exchange limit.}
In the uniform electron gas limit (\(\nabla\rho = 0\), \(\tau = \tau_{\mathrm{UEG}}\)), the short-range DFT exchange enhancement factor, extracted by dividing out both the LDA exchange and the RSH attenuation factor \(f_{\mathrm{SR}}(\rho, \omega)\), must satisfy \(g_x(s{=}0,\,t{=}k_c) = c_{x,0}\), where \(a_{\mathrm{HF}} = 0.15\) is the short-range Hartree--Fock exchange mixing fraction of \(\omega\)B97M-V, giving \(c_{x,0} = 1 - a_{\mathrm{HF}} = 0.85\), and \(k_c = \tfrac{3}{10}(6\pi^2)^{2/3}\) is the UEG value of the kinetic descriptor \(t\). This was tested on synthetic unpolarized UEG grids at four representative densities spanning typical valence-electron regimes. The maximum deviation \(|g_x - c_{x,0}|\) must remain below 0.5\% of true UEG exchange.

\textbf{3. Uniform coordinate scaling of exchange.}
Under uniform coordinate scaling \(\mathbf{r} \to \mathbf{r}/\lambda\), the dimensionless descriptors \(s = |\nabla\rho|/\rho^{4/3}\) and \(t = \tau/\rho^{5/3}\) are invariant. Likewise, the exchange enhancement factor \(g_x(s,t)\) must therefore be unchanged. This was verified on the H\(_2\)O molecular grid at three scaling factors (\(\lambda = 0.5, 2.0, 5.0\)) and comparing the enhancement factor before and after scaling, with the RSH attenuation factor divided out. To achieve scaling invariance, the energy-weighted integrated deviation must fall below \(10^{-4}\)~Ha.

\textbf{4. AE18 grid convergence.}
Atomization energies of the H-Ar are computed on two integration grids: a coarse \((99,\,590)\) and a fine \((250,\,974)\) Lebedev grid. Following the grid-sensitivity protocol established for \(\omega\)B97M-V~\cite{mardirossian2016wB97MV,liang2026coach}, the maximum absolute energy difference across the 18 atomization energies must remain below 0.015~kcal/mol. This threshold ensures that the functional is sufficiently insensitive to integration-grid resolution for practical use, including reliable second-order properties~\cite{sitkiewicz2022reliable}. The behavior of the search when these constraints are not enforced is examined in Appendix~\ref{sec:unconstrained}.

\section{Results}

\subsection{Overall Performance}

Table~\ref{tab:model-comparison} summarizes the best functionals discovered by each LLM backend under each constraint regime. The strongest constrained functional, \texttt{SAFS26-a}, reduces the WRMSD$_{\mathrm{tot}}$ to 3.70 kcal/mol (${\sim}9\%$ improvement over \(\omega\)B97M-V) while satisfying all enforced physical constraints except grid convergence. A parallel grid-constrained search produced \texttt{SAFS26-b}, which passes all four constraints at a WRMSD$_{\mathrm{tot}}$ of 3.83 kcal/mol (${\sim}6\%$ improvement). Both top agent functionals were discovered by the Seed~2.0 backend~\cite{seed2026modelcard}. The GPT~5.4 functionals improve on validation WRMSD but exhibit a widening validation-test gap indicative of benchmark overfitting (Section~\ref{sec:discussion}). The best unconstrained functional(\texttt{SAFS26-x}) achieves competitive performance but fails all four physical constraints (Appendix~\ref{sec:unconstrained}). 

\begin{table}[H]
\centering
\small
\begin{tabular}{lllcc}
\toprule
\textbf{Functional} & \textbf{Model} & \textbf{Constraints} & \textbf{WRMSD}$_{\mathrm{tot}}$ & \textbf{\% Improve.} \\
\midrule
\(\omega\)B97M-V & --- & --- & 4.07 & --- \\
GAS22 & --- & --- & 3.77 & 7.4\% \\
\texttt{SAFS26-x} & Seed~2.0 & None & 3.81 & 6.6\% \\
\texttt{SAFS26-a} & Seed~2.0 & Physical & \textbf{3.70} & \textbf{9.1\%} \\
\texttt{SAFS26-b} & Seed~2.0 & Physical + Grid & 3.83 & 5.9\% \\
\texttt{AFS26-a} & GPT~5.4 & Physical & 3.90 & 4.3\% \\
\texttt{AFS26-b} & GPT~5.4 & Physical + Grid & 3.90 & 4.2\% \\
\bottomrule
\end{tabular}
\caption{WRMSD$_{tot}$ in kcal/mol on MGCDB84 for the best functionals discovered by each LLM backend, grouped by model and constraint regime. The \(\omega\)B97M-V baseline and GAS22~\cite{ma2022evolving} are shown for reference. Full per-split results are given in Table~\ref{tab:full-wrmsd}. SAFS26 = Seed Agentic Functional Search 2026; AFS26 = Agentic Functional Search 2026.}
\label{tab:model-comparison}
\end{table}

\subsection{Per-Category Accuracy}

\definecolor{rc0}{HTML}{FFD3D0}
\definecolor{rc1}{HTML}{C8E6C9}
\definecolor{rc2}{HTML}{FFCDD2}
\definecolor{rc3}{HTML}{FFDDCD}
\definecolor{rc4}{HTML}{FEF9C4}
\definecolor{rc5}{HTML}{DBECC7}
\definecolor{rc6}{HTML}{E6F1C6}
\definecolor{rc7}{HTML}{E5F0C6}
\definecolor{rc8}{HTML}{CCE7C9}
\definecolor{rc9}{HTML}{FFD4D0}
\definecolor{rc10}{HTML}{D3EAC8}
\definecolor{rc11}{HTML}{DFEEC7}
\definecolor{rc12}{HTML}{FFEBC8}
\definecolor{rc13}{HTML}{FFEDC8}
\definecolor{rc14}{HTML}{E7F1C6}
\definecolor{rc15}{HTML}{FFE4CB}
\definecolor{rc16}{HTML}{F6F6C5}
\definecolor{rc17}{HTML}{F1F4C5}

\begin{table}[H]
\centering
\small
\begin{tabular}{l lcccc}
\toprule
\multicolumn{2}{l}{\textbf{Category}} & \(\omega\)B97M-V & GAS22 & \texttt{SAFS26-a} & \texttt{SAFS26-b} \\
\midrule
\ldelim\{{4}{*}[{\rotatebox[origin=c]{90}{\scriptsize Non-covalent}}] & RG10 & \cellcolor{rc0} 2.72 & \cellcolor{rc1} 2.17 & \cellcolor{rc2} 2.76 & \cellcolor{rc3} 2.66 \\
 & NCED & \cellcolor{rc4} 2.25 & \cellcolor{rc1} 2.10 & \cellcolor{rc2} 2.42 & \cellcolor{rc5} 2.15 \\
 & NCEC & \cellcolor{rc2} 6.35 & \cellcolor{rc1} 5.81 & \cellcolor{rc6} 5.96 & \cellcolor{rc7} 5.95 \\
 & NCD & \cellcolor{rc1} 2.36 & \cellcolor{rc8} 2.37 & \cellcolor{rc2} 2.79 & \cellcolor{rc9} 2.75 \\
\midrule
\ldelim\{{5}{*}[{\rotatebox[origin=c]{90}{\scriptsize Covalent}}] & IE & \cellcolor{rc2} 2.98 & \cellcolor{rc1} 2.65 & \cellcolor{rc10} 2.68 & \cellcolor{rc4} 2.81 \\
 & ID & \cellcolor{rc2} 13.25 & \cellcolor{rc11} 10.93 & \cellcolor{rc1} 10.30 & \cellcolor{rc12} 12.23 \\
 & TCE & \cellcolor{rc13} 3.97 & \cellcolor{rc2} 4.16 & \cellcolor{rc1} 3.63 & \cellcolor{rc14} 3.78 \\
 & TCD & \cellcolor{rc14} 1.99 & \cellcolor{rc15} 2.36 & \cellcolor{rc1} 1.76 & \cellcolor{rc2} 2.56 \\
 & BH & \cellcolor{rc2} 8.28 & \cellcolor{rc16} 7.75 & \cellcolor{rc17} 7.71 & \cellcolor{rc1} 7.37 \\
\bottomrule
\end{tabular}
\caption{WRMSD in kcal/mol per MGCDB84 reaction category (validation and test splits only) for \(\omega\)B97M-V, GAS22, and the two strongest discovered functionals. Cells are shaded on a continuous green (best) to red (worst) scale within each row.}
\label{tab:wrmsd-category}
\end{table}

Table~\ref{tab:wrmsd-category} decomposes the aggregate WRMSD improvement into category-level contributions. Functional performance is often a tradeoff between different chemical domains. This is reflected in the per-category performance, which suggests non-uniform improvement, especially between GAS22 and \texttt{SAFS26-a}.
GAS22 leads in the non-covalent categories: non-covalent clusters [easy] (NCEC), non-covalent dimers [easy] (NCED), non-covalent dimers (NCD) [difficult], as well as rare-gas dimers, reflecting well-tuned exchange-correlation behavior in the low-density regime that governs intermolecular binding.
\texttt{SAFS26-a}, by contrast, achieves its largest gains on covalent and thermochemical categories: it substantially outperforms both \(\omega\)B97M-V and GAS22 on the thermochemistry [difficult] (TCD), thermochemistry [easy] (TCE), and isomerizations [difficult] (ID), all categories governed by short-range exchange-correlation in high-density covalent bonding regions. 

\subsection{Search Dynamics}
\label{sec:model_dynamics}

\begin{figure}[H]
\centering
\includegraphics[width=0.7\textwidth]{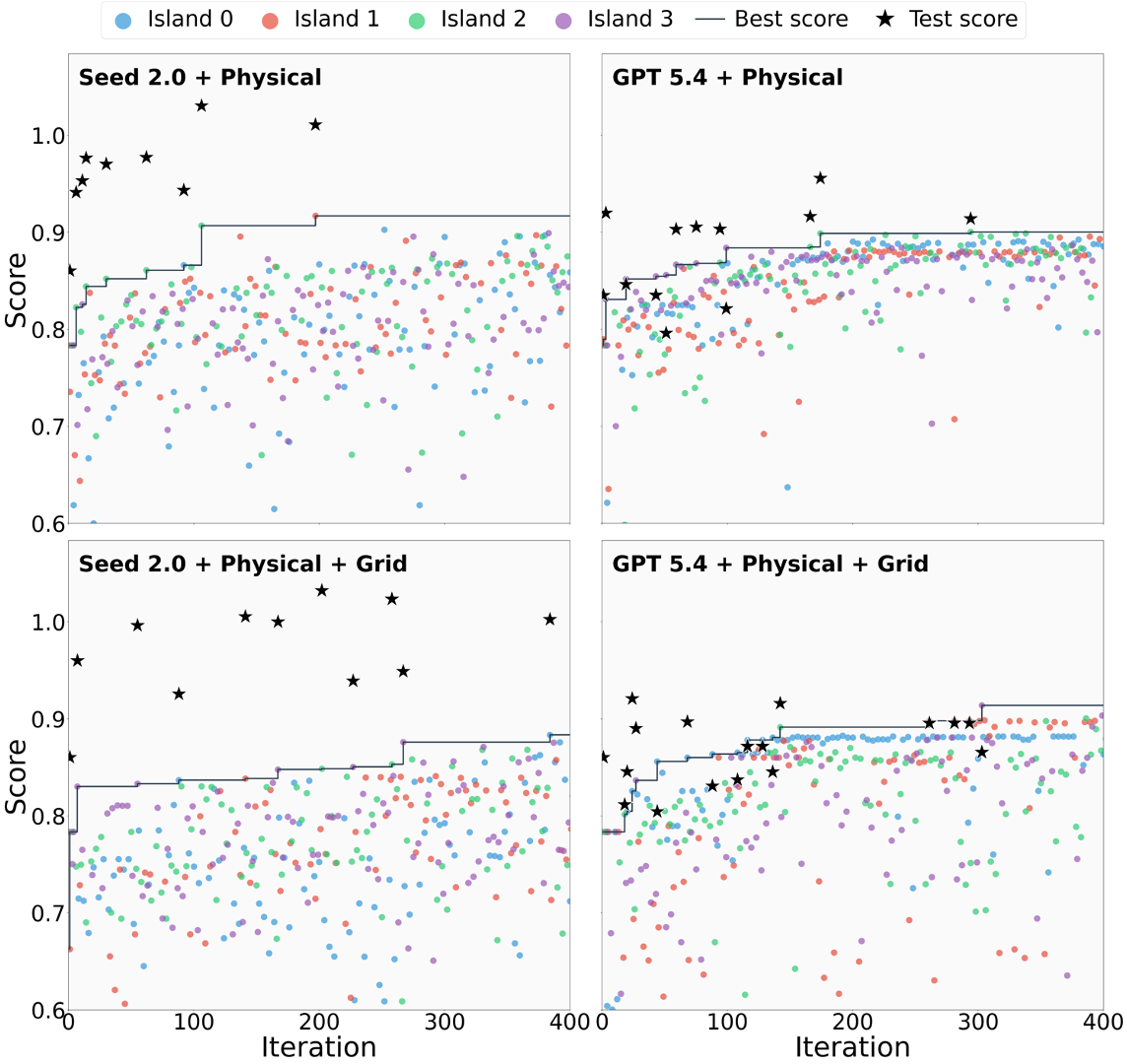}
\caption{Score trajectories over 400 iterations per backend--constraint regime. Dots are per-island candidates, the line is the running best validation score, and stars mark the top scoring test set score (Section~\ref{sec:scoring}). Seed~2.0 explores more broadly with test stars above its validation best, while GPT~5.4 clusters tightly with test oscillating at or below the starting point, signaling validation overfitting.}
\label{fig:score-plot}
\end{figure}

The score trajectories in Figure~\ref{fig:score-plot} hint at a marked difference in proposal style between model backends. Seed~2.0 produces a wide score spread from early iterations, reflecting structurally diverse seed solutions rather than incremental refinement of a single family. This diversity seeds the population with structurally distinct solutions, giving the multi-island architecture richer material for recombination and allowing the cumulative best to saturate sooner. Population-wide plan analysis reveals GPT~5.4's tighter score clustering is a by-product of ignoring the parent solution in favor of the top performing seed. Ablation study in Appendix \ref{sec:ablation-creativity} reveals this proposal pattern also causes overfitting issues. Indeed, although GPT 5.4 performance is tied with the highest score (lowest \(\text{WRMSD}_{val}\)) and displays a significantly higher average score, this is completely offset by their performance on the held out test-set (see Table~\ref{tab:full-wrmsd}).

\subsection{Structural Analysis: \texttt{SAFS26-a}} \label{sec:physical-constrained-func}

The evolutionary trajectory leading to \texttt{SAFS26-a} is detailed in Appendix~\ref{sec:physical-constrained}; below we describe the structural changes in the final functional relative to \(\omega\)B97M-V.

\textbf{Exchange.}
The baseline \(\omega\)B97M-V exchange enhancement is a three-term polynomial in the bounded descriptors \((w, u)\):
\begin{equation}
    g_x = c_{x,0} + c_1 w + c_2 u.
\end{equation}
\texttt{SAFS26-a} retains this polynomial structure but adds one new descriptor: a bounded cross term \(v = st/(1+st+\epsilon)\) coupling gradient and kinetic information

The exchange enhancement becomes
\begin{equation}
    g_x = P_x(w,u) + c_{x,v,0}\,v + c_{x,v,1}\,v^2,
\end{equation}
 where \(P_x(w,u)\) is the baseline \(\omega\)B97M-V polynomial form. 

\textbf{Correlation.}
In \(\omega\)B97M-V, same-spin and opposite-spin correlation are low-order polynomials in \((w,u)\):
\begin{equation}
    g_{\mathrm{\alpha\alpha}} = P_{\mathrm{css}}(w_\alpha,u_\alpha), \qquad
    g_{\mathrm{\beta\beta}} = P_{\mathrm{css}}(w_\beta,u_\beta), \qquad
    g_{\mathrm{\alpha\beta}} = P_{\mathrm{cos}}(w_{\mathrm{avg}},u_{\mathrm{avg}}).
\end{equation}
\texttt{SAFS26-a} replaces these with bounded rational enhancement factors. For each correlation channel, two additional bounded cross descriptors are constructed from the von Barth--Hedin spin-polarization interpolation function \(f_\zeta(\zeta) = [(1+\zeta)^{4/3} + (1-\zeta)^{4/3} - 2] / (2^{4/3} - 2)\)~\cite{vonBarth1972}:
\begin{equation}
    z = \frac{f_\zeta(\zeta) \cdot s \cdot t}{1 + f_\zeta(\zeta) \cdot s \cdot t + \epsilon}, \qquad
    x = \frac{z \cdot s \cdot t}{1 + z \cdot s \cdot t + \epsilon},
\end{equation}
where same-spin channels use per-spin \(s\) and \(t\) while opposite-spin uses spin-averaged values. The enhancement factor is then a bounded rational:
\begin{equation}
    g = \frac{N(w,u,v,z,x)}{D(w,u,v,z,x)},
\end{equation}
with numerator
\begin{equation}
    N = P(w,u) + c_v[0]\,v + c_v[1]\,v^2 + \tanh\!\bigl(c_z[0]\,z + c_z[1]\,z^2\bigr) + c_x[0]\,x + c_x[1]\,x^2,
\end{equation}
and denominator
\begin{equation}
    D = 1 + \sum_i \mathrm{softplus}(d_i)\,\phi_i,
\end{equation}
where the basis \(\{\phi_i\}\) comprises the monomials of \(P_c(w,u)\) together with
\begin{equation}
    \{v,\; v^2,\; z,\; z^2,\; x,\; x^2\}.
\end{equation}
The two channels share this functional form but differ in \(P_c(w,u)\) (each inheriting its monomial powers from the \(\omega\)B97M-V baseline). \(\mathrm{softplus}(\cdot)\) helps ensure non-negative denominator coefficients, suppressing the risk of sign changes. 

The total number of trainable parameters is 50 (compared to 12 in the \(\omega\)B97M-V baseline): 4 exchange parameters (2 polynomial, 2 cross-term \(v\)) and 46 correlation parameters (11 same-spin numerator, 11 same-spin denominator, 12 opposite-spin numerator, 12 opposite-spin denominator). In addition, 6 baseline \(\omega\)B97M-V structural parameters are held fixed: the descriptor mappings (\(\gamma_x\), \(\gamma_{\mathrm{css}}\), \(\gamma_{\mathrm{cos}}\)), and the polynomial power templates.

\subsection{Structural Analysis: \texttt{SAFS26-b}} \label{sec:grid-constrained-func}

The evolutionary trajectory leading to \texttt{SAFS26-b} is detailed in Appendix~\ref{sec:grid-constrained}. The final functional differs from \(\omega\)B97M-V primarily in its correlation channels, with more modest changes to exchange.

\textbf{Exchange.}
Relative to \(\omega\)B97M-V, two modifications were introduced. First, a new descriptor \(v = 1/(1 + \alpha^2)\), where \(\alpha = (\tau - \tau_W)/\tau_\mathrm{UEG}\) is the iso-orbital indicator, was added to the exchange polynomial with two cross terms \(wv\) and \(uv\)~\cite{Perdew2015SCAN}. Second, a multiplicative sigmoid correction was appended:
\begin{equation}
    g_x = P_x(w,u,v)\cdot\bigl(1 + c_{\mathrm{sig},x}\cdot\sigma_x\bigr),
\end{equation}
where \(\sigma(\cdot) = 1/(1+e^{-(\cdot)})\) denotes the logistic sigmoid and \(\sigma_x = \sigma(a_x(s^2 - b_x))\). Although this breaks the UEG constraint slightly, the coefficient values are small, meaning $(1 + c_{\mathrm{sig},x}\cdot\sigma_x) \approx 1$ falls under the threshold. 

\textbf{Same-spin correlation.}
The same-spin enhancement factor combines a polynomial base in \((w,u)\), a sigmoid correction, additive \(u \cdot w\) terms, and a multiplicative rational correction incorporating the Wigner--Seitz radius($r_s$), the \(v\) descriptor, and spin polarization:
\begin{equation}
    g_{\alpha\alpha} = \Bigl[P_{\mathrm{css}}(w_\alpha, u_\alpha) \cdot (1 + c_{\mathrm{sig,ss}} \cdot \sigma_{\mathrm{ss}}^\alpha) + c_1 u_\alpha w_\alpha + c_2 u_\alpha^2 w_\alpha^2\Bigr] \cdot \biggl(1 + \frac{\beta_{\mathrm{ss}} r_{s,\alpha} + c_{v,\mathrm{ss}} r_{s,\alpha} v_\alpha + c_{\zeta,\mathrm{ss}} \zeta_\alpha}{1 + \Gamma_{\mathrm{ss}} r_{s,\alpha}^2}\biggr),
\end{equation}
where \(g_{\beta\beta}\) is defined analogously from \(\beta\)-spin quantities, \(\sigma_{\mathrm{ss}}^\alpha = \sigma(a_{\mathrm{ss}}(t_\alpha - k_c/2))\) gates on the kinetic descriptor, and \(\zeta_\alpha = \zeta(1-\zeta)\), \(\zeta_\beta = -\zeta(1+\zeta)\) are spin-dependent polarization terms ensuring spin symmetry.

\textbf{Opposite-spin correlation.}
The opposite-spin channel received the most extensive modifications. The spin-averaged descriptors used in this channel are computed as \(s_{\mathrm{avg}} = ((s_\alpha^2 + s_\beta^2)/2)^{1/2}\), \(t_{\mathrm{avg}} = 2\,t_\alpha\,t_\beta/(t_\alpha + t_\beta + \epsilon)\), \(v_{\mathrm{avg}} = (v_\alpha + v_\beta)/2\), and \(r_{s,\mathrm{avg}} = (3/(4\pi\rho))^{1/3}\). The base enhancement factor is a polynomial in three descriptors \((w_{\mathrm{avg}}, u_{\mathrm{avg}}, z)\), where \(z = f_\zeta(\zeta)\) is the von Barth--Hedin interpolation function defined above and \(P_{\mathrm{cos}}\) includes 11 terms with cross powers in \(z\) up to quadratic order:
\begin{equation}
    g_{\alpha\beta} = P_{\mathrm{cos}}(w_{\mathrm{avg}}, u_{\mathrm{avg}}, z)  \cdot \biggl(1 + \frac{\beta_{\mathrm{os}} + c_{v,\mathrm{os}} v_{\mathrm{avg}} + c_{\zeta,\mathrm{os}} |\zeta|}{1 + \Gamma_{\mathrm{os}} r_{s,\mathrm{avg}}^2}\biggr),
\end{equation}

The total number of trainable parameters is 19 (compared to 50 for \texttt{SAFS26-a} and 12 for the \(\omega\)B97M-V baseline): 8 scalar correction parameters (2 same-spin additive, 3 \(r_s\)/\(v\)/\(\zeta\) for each of same-spin and opposite-spin) and 11 opposite-spin polynomial coefficients (5 same-spin polynomial coefficients are inherited and frozen from the parent).

\section{Discussion}
\label{sec:discussion}
\subsection{The Value of Structured Evolution Protocol}

A natural question is whether a general-purpose coding agent, operating in a single context window, could achieve an equivalent result if given the same functional-design problem and evaluation pipeline. We argue that two architectural features of the framework, evolutionary memory and population structure, address fundamental limitations of the single-agent setting.

\textbf{Evolutionary memory enables sustained structured search.}
Over 400+ iterations the accumulated evolutionary record (plans, summaries, code, constraint-test outcomes, and scores) grows far beyond the context window of any current LLM. A single-thread agent has limited mechanisms to synthesize lessons across this history: even with tool-augmented context management, the working memory available at each decision point is bounded, making it difficult to maintain a coherent picture of which strategies have been tried, why they failed, and which structural regions remain unexplored (running multiple such agents in parallel does not resolve this: without shared state, each independently rediscovers the same dead ends). Our evolutionary memory \(\mathcal{M}_t\) (Appendix~\ref{sec:evo-memory}) addresses this through three granularities of context: the planner always receives its direct parent's plan, solution, and summarizer diagnosis as immediate context; retrieval tools allow it to selectively query deeper into its lineage history, surfacing ancestor plans and diagnoses that track the broader research trajectory; and a global dead-end list, synthesized across all islands, prevents population-wide repetition of exhausted strategies.

\textbf{Population structure prevents premature convergence.}
A single-thread agent naturally gravitates toward exploiting the current best solution. This is because that is what occupies its context and attention, while suboptimal but structurally distinct solutions are discarded rather than preserved for later refinement. The population structure (Appendix~\ref{sec:pop-structure}) addresses this through two complementary mechanisms. First, the population archive retains structurally diverse solutions that are globally suboptimal but occupy useful regions of the search space; exploration-heavy parent selection  provides repeated chances for these ideas to be refined. Second, island isolation protects nascent ideas through their initially unpromising stages. The decisive fusion that produced \texttt{SAFS26-a} required two qualitatively different ideas---spin-enhanced exchange and rational bounded correlation---to mature independently over multiple iterations on separate islands before being combined at iteration~197 (Appendix~\ref{sec:physical-constrained}). The ablation study in Appendix~\ref{sec:ablation-creativity} demonstrates the consequence of removing these mechanisms. Naively parallelized agent systems can share limited context through shared filesystems or message passing~\cite{anthropic2026agentictrends}, but these coordination mechanisms do not provide the structure needed for reliable ultra-long horizon searches: deciding which suboptimal ideas to preserve, when to revisit them, or how to introduce ideas across lineages.

We should note that significant improvements have been made employing parallel and interacting agents~\cite{anthropic2026agentictrends,lin2026selfdriving,carlini2026compiler}. Nevertheless, these systems have so far been applied primarily to software engineering tasks where correctness is binary and work decomposes naturally into independent sub-tasks. Scientific discovery search poses distinct coordination challenges, including a continuous and deceptive objective landscape, higher risk of benchmark exploitation, and a requirement for strict adherence to domain-specific rules. That said, the rapid progress in multi-agent orchestration is encouraging and we are optimistic that agent architectures will extend their reach from engineering to open-ended scientific optimization.

\subsection{The Risk of Unproductive Shortcuts}

The unconstrained results in Appendix~\ref{sec:unconstrained} illustrate a broader hazard of agentic scientific search: without careful design, optimization pressure will reward unphysical shortcuts over scientifically meaningful improvements. During one unconstrained search, \textit{every} top-scoring functional violated at least three of four basic constraints, and the majority violated all four. The violations are not subtle numerical drift; they represent physics breaking failures (antisymmetric spin terms, improper asymptotic behavior) that were implicitly selected because they lower the training objective. Without explicit constraint enforcement, benchmark exploitation consistently dominates over physically meaningful innovation. This echoes a broader pattern in AI-assisted science: recent work deploying AI for research-level mathematics and physics similarly finds that publishable-quality results required sustained human--AI collaboration, with fully autonomous outputs mostly limited to well-posed problems with verifiable answers~\cite{feng2026autonomous,woodruff2026accelerating}.

Even with all physical constraints enforced, however, a subtler form of benchmark gaming persists at longer search horizons. Inspection of the evolutionary record reveals a recurring pattern: rather than building on their assigned parent solution, agents will source from the current top-performing functional, freeze its existing parameters, and graft small additive corrections on top to eke out marginal score improvements. This behavior becomes especially pronounced when the mechanisms designed to maintain creativity and innovation (multi-island isolation and exploration-heavy parent selection), are removed from the evolution architecture; the ablation study in Appendix~\ref{sec:ablation-creativity} documents the most extreme instance, a functional that froze 61 parameters from its parent and trained only 4, achieving a record-low validation WRMSD while generalizing substantially worse than the \(\omega\)B97M-V baseline on the test set; instead of exploring qualitatively different functional forms, the search collapsed into repeated micro-perturbations of a single elite solution. Although each perturbation is trained on the training split, it is evolved on its performance on the validation split. Over many generations, the cumulative effect of hundreds of such incremental additions is to progressively leak the validation set. The overfitting signature is unambiguous and gets worse with time: when the grid-constrained search is extended from 400 to 800 iterations, the validation WRMSD continues to improve slightly, but the test WRMSD degrades substantially, a hallmark of data leakage rather than genuine generalization. This underscores that physical constraints alone are insufficient to prevent all forms of benchmark exploitation.

\subsection{Conclusion}
This work demonstrates that an agent-guided evolutionary search can discover exchange--correlation density functionals that improve over a strong human-designed baseline while respecting enforced constraints. Starting from \(\omega\)B97M-V, the search produced \texttt{SAFS26-a}, which reduces the WRMSD$_{tot}$ ${\sim}9\%$ and satisfies three of four constraints, and \texttt{SAFS26-b}, which passes all four constraints with a ${\sim}6\%$ WRMSD$_{tot}$ improvement. These results reveal three properties of agent-driven search that are relevant beyond DFT. First, structured evolutionary memory  enables agents to sustain coherent searches over hundreds of iterations without collapsing into repetition or blowing up context, a capability that single-conversation coding agents lack because their context is bounded and eventually discarded or compressed. Second, population diversity is essential, and multiple steps were needed to maintain it: the multi-island topology, high exploration, and large population archive maintain a pool of structurally distinct ``stepping-stone'' solutions that were individually suboptimal but collectively enabled the cross-lineage fusions responsible for the largest score improvements. Third, creative high-variance proposal generation is a strength rather than a liability in population-based search; over a long span, diversity of ideas matters more than per-iteration reliability. This stands in contrast to recent LLM post-training paradigms centered on verifiable domains such as mathematics, coding, and formal reasoning, where reward signals derive from objectively correct solutions~\cite{guo2025deepseekr1,wei2025swerl}. Our results suggest that in discovery-oriented settings, preserving high-variance exploratory proposal generation may be more valuable than maximizing short-horizon correctness.

Equally important is the central cautionary finding: the same search machinery, without physical constraints, reliably produces functionals that game the benchmark through unproductive mechanisms (Appendix ~\ref{sec:unconstrained}). Every top-scoring unconstrained functional violated at least three of four basic constraints. The physical constraints imposed by domain expertise are not incidental to the result: they are what separates scientifically meaningful improvement from benchmark exploitation. Unless these guardrails are carefully enforced or the task itself is very well defined, the trustworthy role of agent-based scientific agents remains constrained search under human oversight rather than fully autonomous research.

\section{Future Directions}

As previously stated, validation-set leakage remains a problem for these models. The most direct mitigation of this is to train and score on a larger, more chemically diverse benchmark~\cite{liang2025gold}; a richer data distribution raises the statistical cost of the incremental micro-perturbations that drove leakage here, making exploitation of any narrow subset less rewarding. The increased breadth and complexity of such a corpus---spanning transition-metal energetics, barrier heights, and electric-field responses in addition to the thermochemistry and noncovalent interactions of MGCDB84---would simultaneously constitute a more stringent test of the agent's capacity for genuine chemical innovation, requiring discovered functionals to generalize across a broader swath of chemical space. This is far from the only mitigation factor, penalization scoring based on problematic behavior (i.e. perturbative additions onto frozen parameters) or semantic similarity may also afford the intended effect. A natural progression of this pipeline would be to include a judge or rubric agent to systematically grade trial functionals on these more nebulous metrics. 

Second, the results reported here rely on non-self-consistent evaluation using electron densities held fixed at the \(\omega\)B97M-V solution. Computing SCF-converged energies for the discovered functionals is a necessary follow-up to confirm that the observed improvements persist once the density is allowed to relax. More broadly, incorporating periodic rounds of SCF convergence into the evolutionary search itself (in the spirit of established protocols~\cite{liang2026coach}) would ensure that the frozen-density approximation remains valid as the functional form drifts further from the baseline, preventing the search from inadvertently optimizing for an increasingly inaccurate surrogate.

Finally, the functional-evolution task is naturally suited to LLM post-training through reinforcement learning: the symbolic functional form defines a state, the LLM-proposed modification is an action, and the WRMSD score provides a continuous reward. Recent work on test-time RL~\cite{yuksekgonul2026learningdiscovertesttime} has shown that updating model weights on a single problem (rather than prompting a frozen LLM) can discover stronger solutions by internalizing lessons from the search trajectory, and applying such an approach to functional evolution could complement or replace the prompt-based evolutionary memory used here.

\bibliographystyle{unsrt}
\bibliography{references}

\appendix

\section{Full Train/Validation/Test WRMSD}
\label{sec:full-wrmsd}

\begin{table}[H]
\centering
\small
\begin{tabular}{lccc}
\toprule
\textbf{Functional} & \textbf{WRMSD}$_{\mathrm{train}}$ & \textbf{WRMSD}$_{\mathrm{val}}$ & \textbf{WRMSD}$_{\mathrm{test}}$ \\
\midrule
\(\omega\)B97M-V & 3.32 & 4.02 & 3.64 \\ 
GAS22 & 3.27 & 3.74 & \textbf{3.19} \\ 
\texttt{SAFS26-x} & 2.91 & 3.95 & 3.49 \\ 
\texttt{SAFS26-a} & 2.76 & \textbf{3.65} & 3.41 \\ 
\texttt{SAFS26-b} & 2.75 & 3.78 & 3.44 \\ 
\texttt{AFS26-a} & 2.81 & 3.74 & 3.78 \\ 
\texttt{AFS26-b} & \textbf{2.65} & 3.65 & 3.96 \\ 
\bottomrule
\end{tabular}
\caption{Per-split WRMSD in kcal/mol on MGCDB84 for each functional. Bold entries indicate the best value in each column among the discovered functionals. The \(\omega\)B97M-V baseline and GAS22~\cite{ma2022evolving} are shown for reference.}
\label{tab:full-wrmsd}
\end{table}

\section{Evolutionary Framework}
\label{sec:evo-framework}

\subsection{Planner: lineage-aware hypothesis generation}
\label{sec:planner}
The planner is the central component of the agentic framework. Given a parent solution \(s_t\), it generates a natural-language blueprint \(b\) conditioned on the evolutionary memory \(\mathcal{M}_t\):
\begin{equation}
b \sim \pi_\theta\!\bigl(b \mid s_t,\;\mathcal{M}_t,\;\mathcal{I}_{\mathrm{plan}}\bigr),
\end{equation}
where \(\mathcal{I}_{\mathrm{plan}}\) is the planner system prompt. The construction and delivery of \(\mathcal{M}_t\) is described in Section~\ref{sec:evo-memory}.

In practice, the planner's blueprints are detailed enough to specify the mathematical form of the proposed change, the initialization of new parameters, and the rationale for why the change should improve accuracy.

\subsection{Executor: implementation and verification}
\label{sec:executor}
The executor translates the planner's blueprint into executable Python code implementing the new functional form:
\begin{equation}
s' \sim \pi_\theta\!\bigl(s' \mid b,\;s_t,\;\mathcal{I}_{\mathrm{exec}}\bigr).
\end{equation}
Its value lies not in proposing new ideas but in implementation quality: correctly translating mathematical specifications into executable computational components, handling bounded descriptors, and ensuring numerical stability. Before the candidate is sent to the full training-and-evaluation pipeline, the executor runs a local verification loop that catches syntax errors and import failures, preventing low-quality candidates from consuming expensive evaluation resources.

When the plan is ambiguous (for example, specifying unpolarized formulas for a spin-resolved functional), the executor must interpret the physics and select the correct spin-polarized variants. The trajectories show instances where the executor also debugged numerical instabilities (e.g., replacing manual polynomial constructions with stable JAX built-ins) through multiple iterative attempts before producing a working implementation.

\subsection{Summarizer: causal analysis for future iterations}
\label{sec:summarizer}
After evaluation, the summarizer compares the planner's intent with the actual outcome and generates a structured insight:
\begin{equation}
z \sim \pi_\theta\!\bigl(z \mid s',\;R(s'),\;b,\;\mathcal{I}_{\mathrm{sum}}\bigr), \qquad \mathcal{M}_{t+1} \leftarrow \mathcal{M}_t \cup \{z\}.
\end{equation}
Each summary contains an executive summary, data-driven findings (score deltas, sibling rankings), a root-cause analysis of success or failure, a key insight for future iterations, and identified risks. Critically, the summarizer is instructed to report facts and diagnose causes, not to suggest strategies: that responsibility belongs to the planner.

This separation ensures that the accumulated memory is a reliable factual record rather than a set of potentially contradictory recommendations.

\subsection{Population structure and selection}
\label{sec:pop-structure}
The population is managed by a multi-island evolutionary model with two components:
\begin{itemize}[leftmargin=*]
\item \textbf{Multi-island topology.} The population is partitioned into \(K\) islands (four in this work), each evolving independently. Different islands can develop distinct functional-form ``species'' in isolation. At regular intervals, the top elites from each island are migrated to neighboring islands.

\item \textbf{Elite archive and exploitation.} Each island maintains a capped set of the \(k\) highest-scoring solutions (elites). Parent selection draws uniformly at random from this elite set, concentrating compute on the most promising regions of the search space. The exploration rate \(\epsilon\) overrides this and selects a uniformly random solution from the full population, providing the exploration counterpart.
\end{itemize}

\subsection{Evolutionary memory}
\label{sec:evo-memory}
The evolutionary memory \(\mathcal{M}_t\) reaches the planner's context through two complementary channels:
\begin{itemize}[leftmargin=*]
\item \textbf{Per-lineage retrieval.} The planner is equipped with database tools to query its own lineage history: the plans that created each ancestor and the summarizer outputs that recorded what worked or failed. This provides \textit{intent tracking} (understanding the research trajectory of prior generations) and local \textit{course correction} (avoiding strategies already tried within its lineage).

\item \textbf{Global memory synthesis.} Before each planning step, an LLM synthesizer scans the full iteration history across all islands and injects a list of \textit{exhausted dead ends} directly into the planner's context. A strategy is declared exhausted when it has been attempted many times across the population, never improved the score, and its failures share a common root cause. This global channel is necessary because reconstructing a cross-island picture through sequential tool calls would exceed practical context limits.
\end{itemize}

\section{Task Definition}
\label{sec:task-definition}

The following is the exact task description passed to the LLM planner and executor at every iteration of the evolutionary search.

\begin{tcolorbox}[colback=gray!5,colframe=gray!75!black,boxrule=0.5pt,arc=2mm,title=Task Definition]
\small
You are a senior computational chemist implementing a new local (short-range exchange + correlation) functional for a range-separated hybrid meta-GGA.

\medskip\noindent\textbf{Goal.}
Create a more physically accurate DFT functional, as measured by the WRMSD on a comprehensive thermochemistry benchmark dataset.
Aim to innovate beyond existing architectures rather than replicating them.

\medskip\noindent\textbf{What You're Building.}
The \textbf{local/semi-local part} of a range-separated hybrid meta-GGA functional.
Long-range HF exchange ($\omega{=}0.3$, $c_x{=}0.15$) and VV10 dispersion ($b{=}6$, $C{=}0.01$) are \textbf{fixed} and cannot be modified or trained.
You implement: attenuated short-range exchange + local correlation.

\medskip\noindent\textbf{Execution Notes.}
\begin{itemize}[leftmargin=*,itemsep=0pt,topsep=2pt]
  \item You \textbf{must} view \texttt{base.py} \textit{before} writing any code.
  \item Re-use existing utilities where appropriate, but you are not limited to them.
  \item Your solution file must contain only \textbf{one} functional class; full training runs automatically.
\end{itemize}

\medskip\noindent\textbf{Available Utilities.}
\begin{itemize}[leftmargin=*,itemsep=0pt,topsep=2pt]
  \item \texttt{SeparatedPxExcFunc}: base class; exchange defined for polarized spins, correlation handles spin internally.
  \item \texttt{spinsum\_and\_zeta(rho, drho, tau)} $\to$ \texttt{(rho\_total, zeta, drho\_total, tau\_total)}
  \item \texttt{spin\_scaling\_exchange\_polarized(fn)}: decorator for exchange on single spin channel.
  \item \texttt{spin\_decompose\_correlation(fn)}: wraps correlation to return \texttt{(ec\_aa, ec\_bb, ec\_ab)}.
  \item \texttt{spin\_fz($\zeta$)}: $[(1{+}\zeta)^{4/3}+(1{-}\zeta)^{4/3}-2]\,/\,(2^{4/3}-2)$;\quad \texttt{spin\_phi}, \texttt{spin\_ds}, \texttt{spin\_dx}: analogous spin-scaling functions.
  \item \texttt{gga\_s\_from\_drho}: $|\nabla\rho|/\rho^{4/3}$;\quad \texttt{mgga\_t\_from\_tau}: $\tau/\rho^{5/3}$;\quad \texttt{fermi\_kf}: Fermi wavevector $k_F$.
  \item \texttt{lda\_x\_slater\_pol}: Slater exchange (single spin);\quad \texttt{lda\_c\_pw}: PW92 correlation;\quad \texttt{pw92\_g}, \texttt{pw92\_g\_deriv}: PW92 $g(r_s)$ and its derivative.
  \item \texttt{rsh\_exchange\_factor(rho, omega, polarized)}: short-range attenuation factor ($\omega{=}0.3$ fixed).
\end{itemize}
\end{tcolorbox}

\section{Physical Constraint Search: Discovery of \texttt{SAFS26-a}} \label{sec:physical-constrained}
The evolutionary record does not show smooth monotonic improvement. Instead, it reveals a small number of reusable functional-form ideas that appeared on separate islands, stalled independently, and were ultimately combined across lineages. The best functional, \texttt{SAFS26-a}, emerged at the intersection of two independently evolved branches.

\subsection{Discovery trajectory (iterations 14--197)} \label{sec:physical-constrained-trajectory}
The direct lineage of \texttt{SAFS26-a} runs through two independently evolved branches on Island~2, that migrated across islands and ultimately fused at iteration~197.

\textbf{Iterations 14--97: Spin-exchange lineage.}
The first lineage began on Island~2 and proceeded through three generations of incremental descriptor additions to the \(\omega\)B97M-V baseline. The first descendant, \texttt{ExtendedWB97MV} (iteration~14), introduced a bounded cross descriptor \(v = st/(1+st+\epsilon)\), to all three enhancement factor channels (exchange, same-spin correlation, opposite-spin correlation). This single addition reduced the performance non-negligibly (WRMSD$_{val}$ 4.40 \(\to\) 4.09) kcal/mol. 

\begin{tcolorbox}[colback=gray!5,colframe=gray!75!black,boxrule=0.5pt,arc=2mm]
\small
\textbf{Iteration 14 summary.} ``Bounded cross-terms between \(s\) and \(t\) can add accuracy without breaking constraints. The new \(v\) descriptor captures gradient-kinetic interactions that the independent \(w\) and \(u\) descriptors cannot represent.''
\end{tcolorbox}

The second descendant, \texttt{SpinEnhancedExtendedWB97MV} (iteration~30), added spin-dependent \(z\) descriptors to the correlation channels only, constructed as \(z = f(\zeta) \cdot s \cdot t / (1 + f(\zeta) \cdot s \cdot t + \epsilon)\) where \(f(\zeta) = [(1+\zeta)^{4/3} + (1-\zeta)^{4/3} - 2] / (2^{4/3} - 2)\) is the von Barth--Hedin spin-polarization interpolation function, bounded in \([0,1]\) with unpolarized and polarized cases returning \(f(0)=0\) and \(f(\pm 1)=1\). This yielded a modest improvement (WRMSD$_{val}$ 4.09 \(\to\) 4.05) kcal/mol. The third descendant, \texttt{SpinExEnhancedWB97MV} (iteration~97), extended the same idea to exchange, adding a spin-dependent \(z_x\) descriptor. The gain was negligible: 

\begin{tcolorbox}[colback=gray!5,colframe=gray!75!black,boxrule=0.5pt,arc=2mm]
\small
\textbf{Iteration 97 summary.} ``The negligible improvement is attributable to high quality execution of a plan with overestimated expected gains: the short-range exchange channel was already sufficiently optimized in prior iterations, so adding small spin-dependent cross terms provided no meaningful accuracy benefit.''
\end{tcolorbox}

\noindent By iteration~97, this lineage had reached a form-level plateau at WRMSD$_{val}$~\(\approx\)~0.006454.

\textbf{Iterations 18--106: Rational-correlation lineage.}
A second, independent lineage independently made progress using a different approach. Its first descendant, \texttt{WB97MV\_Rational} (iteration~18), attempted to replace both exchange and correlation enhancement factors with rational forms \(P(w,u)/Q(w,u)\). The raw WRMSD$_{val}$ improved (4.40 \(\to\) 4.17) kcal/mol, but the trainable denominator terms shifted the exchange enhancement factor away from the required UEG value, causing a constraint failure: 

\begin{tcolorbox}[colback=gray!5,colframe=gray!75!black,boxrule=0.5pt,arc=2mm]
\small
\textbf{Iteration 18 summary.} ``Raw valid WRMSD improved, but UEG exchange constraint failed. The trainable denominator broke the UEG guarantee. The constant term of the denominator must be fixed to 1 to preserve \(g_x = 0.85\) at the UEG limit.''
\end{tcolorbox}

\noindent This failure was instructive: the summarizer's causal diagnosis (that the denominator's leading term must be fixed, not trainable) was preserved in evolutionary memory and informed subsequent designs. The next descendant, \texttt{TanhBoundedSpinCrossEnhancedWB97MV} (iteration~82), abandoned the rational exchange form entirely and instead added tanh-bounded spin descriptors and a new cross descriptor \(x = zv/(1+zv+\epsilon)\) to the correlation channels while keeping exchange as a standard polynomial. All four constraints passed, and the performance from its parent improved (WRMSD$_{val}$ 4.17 \(\to\) 4.08) kcal/mol. 

The third descendant, \texttt{RationalBoundedCrossEnhancedWB97MV} (iteration~106), reintroduced rational forms, but only to correlation, and with a critical fix: all denominator coefficients were wrapped in \(\mathrm{softplus}(x) = \log(1+e^x)\), guaranteeing non-negative contributions and a denominator \(D \geq 1\) everywhere.

\begin{tcolorbox}[colback=gray!5,colframe=gray!75!black,boxrule=0.5pt,arc=2mm]
\small
\textbf{Iteration 106 plan.} ``Replace correlation enhancement factors with bounded rational form. The numerator retains all existing terms; the denominator uses softplus-wrapped coefficients so that \(D \geq 1\) always, eliminating division by zero risk. Exchange is unchanged.''
\end{tcolorbox}

\noindent This produced a WRMSD$_{val}$ of 3.81 kcal/mol, the highest score on any single island before the fusion event. The key design insight, learned from the iteration~18 failure, was that rational forms should be applied to correlation only, with exchange left as a simple polynomial to preserve the UEG constraint algebraically. 

\textbf{Iteration 197: fusion as the decisive step.}
By iteration~197, the two branches were joined through migration across islands: the spin-exchange lineage's \texttt{SpinExEnhancedWB97MV} resided on Island~1, and the rational-correlation lineage's \texttt{RationalBoundedCrossEnhancedWB97MV} migrated to from Island~2 to Island~1. The planner recognized that the two lineages had complementary strengths (one in exchange, the other in correlation) and proposed fusing them:

\begin{tcolorbox}[colback=gray!5,colframe=gray!75!black,boxrule=0.5pt,arc=2mm]
\small
\textbf{Iteration 197 plan.} ``Fusion of the current parent's spin-enhanced exchange channel with the top-performing rational bounded correlation channel from island 1's best solution, delivering combined improvements from both independent high-performing lineages. [\ldots] Initializes exactly to the performance level of the top island 1 solution when exchange spin parameters are set to 0, eliminating all regression risk.''
\end{tcolorbox}

The fusion produced \texttt{FusionRationalSpinEnhancedWB97MV}(\texttt{SAFS26-a}): exchange from the spin-exchange lineage (with the spin-dependent \(z_x\) descriptor) combined with correlation from the rational-correlation lineage (rational bounded forms with softplus denominators and tanh-bounded spin cross terms). The result was a WRMSD$_{val}$ of 3.65 kcal/mol, a significant score improvement over the direct parent: 

\begin{tcolorbox}[colback=gray!5,colframe=gray!75!black,boxrule=0.5pt,arc=2mm]
\small
\textbf{Iteration 197 summary.} ``Score improved substantially. [\ldots] Fusion of independently optimized, complementary functional components across separate lineages delivers far larger performance gains than incremental descriptor additions to already optimized single channels.''
\end{tcolorbox}

\noindent The functional remained the global best through the entire remaining 200 iterations of the search (iterations 198--400), with no subsequent modification able to improve upon it.

\textbf{Agent Trajectory Caveat} Iteration~97 describes an additional exchange spin-dependent cross-term descriptor. This term is excluded from Section \ref{sec:physical-constrained-func} because an oversight led to the effects of the spin becoming nullified. Because the exchange enhancement factor is evaluated separately on each spin channel's (scalar) density, the spin-resolution needed to compute \((\zeta)\) collapses by the time the functional is called, nullifying explicit \((\zeta)\)-dependence and effectively yielding (\(z_x\approx v\)).

\section{Grid-Constrained Search: Discovery of a Fully Compliant Functional}
\label{sec:grid-constrained}

The search described in Section \ref{sec:physical-constrained} enforced spin symmetry, UEG exchange limit, uniform coordinate scaling but did not penalize AE18 grid sensitivity during evolution. A parallel search was conducted with AE18 grid convergence added as a fourth enforced constraint.

\subsection{Discovery trajectory (iterations 8--384)}
\label{sec:grid-trajectory}

The direct parent chain of \texttt{SAFS26-b} lies on Island~0, but the two largest score jumps each drew on solutions from other islands: \texttt{SigmoidRegimeTunableSwitch} (Island~1, iteration~141, WRMSD$_{val}$ 4.12 kcal/mol) and \texttt{EvolveSRMGGA\_ZetaAugmented\_Optimized} (Island~3, iteration~267, WRMSD$_{val}$ 3.94 kcal/mol). One marginal intermediate generation (iteration~288, \(r_s \times v\) cross terms) is omitted from the narrative below. 

\textbf{Iterations 8--80: alpha descriptor, UEG failure, and descriptor partitioning.}
The first descendant, \texttt{WB97MV\_AlphaDescriptor} (iteration~8), introduced an iso-orbital indicator descriptor \(v = 1/(1+\alpha^2)\), (where \(\alpha = (\tau - \tau_W)/\tau_{\mathrm{UEG}}\) is the iso-orbital indicator) to all three enhancement-factor channels. The iso-orbital indicator \(\alpha\) is the same quantity whose grid sensitivity problems are well documented in the SCAN functional~\cite{wheeler2010grid}; however, the functional does not use \(\alpha\) directly in interpolation functions with sharp features as SCAN does. Instead, it maps \(\alpha\) through a smooth inverse decay \(v = 1/(1+\alpha^2)\), which saturates to zero for large \(\alpha\) and is clipped to \([0.01,\,1.0]\). This bounded mapping suppresses the effect of large grid-dependent fluctuations in \(\alpha\) that occur in low-density tail regions. 

The validation WRMSD$_{val}$ improved relative to the parent seed (4.40 \(\to\) 4.34) kcal/mol, but the \(v\) descriptor in exchange shifted the enhancement factor away from the UEG limit: 

\begin{tcolorbox}[colback=gray!5,colframe=gray!75!black,boxrule=0.5pt,arc=2mm]
\small
\textbf{Iteration 8 summary.} ``UEG Exchange constraint failed with max \(|g_x - 0.85| = 1.06 \times 10^{-2}\). [\ldots] Expanding the functional form with new descriptors without explicitly fixing their limiting values for standard physical constraint cases leads to avoidable constraint violations even when raw performance improves.''
\end{tcolorbox}

\noindent The next descendant, \texttt{CorrelationOnlyVDescriptor} (iteration~80), proposed removing \(v\) from exchange while retaining it in correlation, directly citing the parent's failure:

\begin{tcolorbox}[colback=gray!5,colframe=gray!75!black,boxrule=0.5pt,arc=2mm]
\small
\textbf{Iteration 80 plan.} ``Fixes the parent's UEG Exchange constraint violation by eliminating \(v\) terms from the exchange channel, which were the root cause of \(g_x\) deviating from the required 0.85 value at the UEG limit. Retains all \(v\) descriptor terms in correlation channels, which were proven in the parent iteration to deliver measurable improvement in raw thermochemistry prediction accuracy.''
\end{tcolorbox}

\noindent All four constraints now passed. The raw WRMSD$_{val}$ was slightly worse (4.54 kcal/mol), but the score improved because the UEG penalty was eliminated. No subsequent generation reintroduced descriptors to the exchange channel. 

\textbf{Iteration 192: absorbing the sigmoid regime-switch architecture from Island 1.}
The largest single-step gain in the lineage came at iteration~192, where the planner absorbed the sigmoid regime-switch architecture from \texttt{SigmoidRegimeTunableSwitch}, a solution on Island~1 with WRMSD$_{val}$ of 4.12 kcal/mol. That solution had independently evolved tunable sigmoid switches for exchange and additive \(u \cdot w\) correction terms for same-spin correlation. The iteration~192 planner combined this architecture with \(r_s\)-dependent rational correction terms for correlation: 

\begin{tcolorbox}[colback=gray!5,colframe=gray!75!black,boxrule=0.5pt,arc=2mm]
\small
\textbf{Iteration 192 plan.} ``Combining the highest-performing sigmoid regime switch architecture (score 0.8385) with validated \(r_s\)-dependent rational correction terms for correlation channels [\ldots] Merges two independently validated performance improvements: tunable exchange regime switches and \(r_s\)-dependent correlation corrections, targeting complementary error sources (high-gradient bonding regions and low-density correlation effects) for synergistic improvement.''
\end{tcolorbox}

\noindent The exchange enhancement factor was augmented with a multiplicative sigmoid correction \((1 + c_{\mathrm{sig}} \cdot \sigma(a_x(s^2 - b_x)))\) and a trainable switch function \(f_{\mathrm{switch}}(u)\). Same-spin correlation received an \(r_s\)-dependent multiplicative factor \((1 + \beta_{\mathrm{ss}} r_s / (1 + \gamma_{\mathrm{ss}} r_s^2))\), where \(\beta_{\mathrm{ss}}\) is a trainable coefficient controlling the linear-\(r_s\) correction strength and \(\gamma_{\mathrm{ss}} = 0.1\) is a fixed regularization constant whose sole role is to damp the correction at large \(r_s\), preventing divergence in the low-density limit (it is unrelated to the descriptor-mapping parameters \(\gamma_x\), \(\gamma_{\mathrm{css}}\), \(\gamma_{\mathrm{cos}}\)); opposite-spin correlation received an analogous term. The score improved substantially (WRMSD$_{val}$ 4.54 \(\to\) 4.13) kcal/mol: 

\begin{tcolorbox}[colback=gray!5,colframe=gray!75!black,boxrule=0.5pt,arc=2mm]
\small
\textbf{Iteration 192 summary.} ``Score improved significantly. [\ldots] Combining pre-validated, complementary architectural modifications targeting separate electronic regimes delivers measurable performance gains without increasing constraint violation risk.''
\end{tcolorbox}

\textbf{Iteration 332: spin-polarization corrections.}
After a marginal intermediate generation that added \(r_s \times v\) cross terms (iteration~288), the iteration~332 planner identified spin polarization as an unexploited structural dimension:

\begin{tcolorbox}[colback=gray!5,colframe=gray!75!black,boxrule=0.5pt,arc=2mm]
\small
\textbf{Iteration 332 plan.} ``Targets the key unexplored structural gap in the parent architecture: existing correlation terms do not account for spin polarization dependence, which is a known source of error for spin-polarized systems in thermochemistry benchmarks. [\ldots] Leaves the exchange channel completely unchanged to preserve the parent's already good UEG constraint performance.''
\end{tcolorbox}

\noindent Spin-polarization (\(\zeta\))-dependent terms were added to the correlation corrections: a \(\zeta(1-\zeta)\) term for same-spin channels and an \(|\zeta|\) term for opposite-spin (WRMSD$_{val}$ 4.12 \(\to\) 4.04) kcal/mol: 

\begin{tcolorbox}[colback=gray!5,colframe=gray!75!black,boxrule=0.5pt,arc=2mm]
\small
\textbf{Iteration 332 summary.} ``Score improved. [\ldots] The trained \(c_{\mathrm{ss},\zeta} = -0.770\) indicates that spin-polarization corrections to same-spin correlation carry substantial weight.''
\end{tcolorbox}

\textbf{Iteration 384: absorbing zeta polynomial cross terms from Island 3.}
The final and second-largest gain came from absorbing the opposite-spin zeta polynomial architecture from \texttt{EvolveSRMGGA\_ZetaAugmented\_Optimized}, a solution on Island~3. That solution had independently developed an 11-term polynomial in \((w_{\mathrm{avg}}, u_{\mathrm{avg}}, z)\) for opposite-spin correlation, where \(z = f_\zeta(\zeta)\) is the von Barth--Hedin interpolation function defined above:

\begin{tcolorbox}[colback=gray!5,colframe=gray!75!black,boxrule=0.5pt,arc=2mm]
\small
\textbf{Iteration 384 plan.} ``Extending the validated parent by merging the zeta polynomial cross terms from the global top-performing \texttt{EvolveSRMGGA\_ZetaAugmented\_Optimized} architecture exclusively into the opposite-spin correlation channel, with no changes to exchange or same-spin correlation channels, plus explicit \texttt{sig\_coeff\_os} clamping. [\ldots] Only modifies the opposite-spin correlation channel, which is the highest-impact underoptimized channel per past iteration results.''
\end{tcolorbox}

\noindent The opposite-spin polynomial basis was extended to include \(z\) as a third descriptor. The \texttt{sig\_coeff\_os} parameter was explicitly clamped to \([0, 2]\). The resulting functional, \texttt{WB97MV\_RegimeSwitch\_RSVZetaPolynomial}(\texttt{SAFS26-b}), produced the best score in the lineage (WRMSD$_{val}$ 4.04 \(\to\) 3.78) kcal/mol: 

\begin{tcolorbox}[colback=gray!5,colframe=gray!75!black,boxrule=0.5pt,arc=2mm]
\small
\textbf{Iteration 384 summary.} ``Score improved. [\ldots] Merging validated features from top-performing architectures exclusively into specific underoptimized channels delivers incremental performance gains while maintaining full constraint compliance with zero regression risk when new parameters are initialized to zero.''
\end{tcolorbox}

\textbf{Agent Trajectory Caveat} Iteration 384  describes additional clipped sigmoid term added to the same spin correlation: \((1 + \mathrm{clamp}(c_{\mathrm{sig,os}}, 0, 2) \cdot \sigma(a_{\mathrm{os}} \cdot s_{\mathrm{avg}}^2))\). This term is excluded from Section \ref{sec:grid-constrained-func} because \(c_{\mathrm{sig,os}}\) optimized below the clipping range, nullifying its effects. 

\section{Ablation Study: Dangers of an Unconstrained Search}
\label{sec:unconstrained}

As noted in Table~\ref{tab:model-comparison}, the top unconstrained functional (\texttt{SAFS26-x}) achieves a WRMSD$_{val}$ comparable to the constrained search while failing all four physical constraints. This subsection examines \textit{how} the optimizer exploits unphysical degrees of freedom.

\noindent Inspection of the implementation reveals the specific mechanisms by which the optimizer exploited unphysical degrees of freedom.

\textbf{1. Spin symmetry violation (3.5~mHa; tolerance 0.01~mHa).}
The opposite-spin correlation enhancement factor includes an additive polarization term:
\begin{codebox}
zeta = (rho_s[0] - rho_s[1]) / (rho_s.sum(0) + EPS)
g_ab = g_ab + alpha_cos * zeta
\end{codebox}
The optimizer converged to \(\texttt{alpha\_cos} = -0.174\), which is antisymmetric under spin exchange: swapping \(\rho_\alpha \leftrightarrow \rho_\beta\) flips the sign of \(\zeta\), directly breaking the invariance \(E_{\mathrm{xc}}[\rho_\alpha,\rho_\beta] = E_{\mathrm{xc}}[\rho_\beta,\rho_\alpha]\). The optimizer retained this term because the antisymmetric correction lowers training loss on open-shell systems like O\(_2\).

\textbf{2. UEG exchange limit violation (1.67\%; tolerance 0.5\%).}
The exchange enhancement factor passes the descriptors \((w, u)\) along with a normalized Laplacian through a learned two-unit sigmoid layer before constructing \(g_x\):
\begin{codebox}
feats = jnp.stack([w, u, lapl_norm], axis=-1)
hidden = jax.nn.sigmoid(feats @ W_x.T + b_x)
g_x = c_x_0 + (hidden * c_x_hidden).sum(-1) * (
    1 + c_train_base[0]*w + c_train_base[1]*u + ...)
\end{codebox}
In the uniform electron gas (\(s=0\), \(\tau=\tau_{\mathrm{UEG}}\)), the descriptors evaluate to \(w=0\), \(u=0\), \(\mathrm{lapl\_norm}=0\), and the sigmoid units produce \(\sigma(b_x) \neq 0\). The resulting hidden-layer residual shifts \(g_x\) away from the required value \(c_{x,0}=0.85\) by 1.67\%. In the baseline \(\omega\)B97M-V, exchange is a simple polynomial that evaluates to \(c_{x,0}\) exactly when \(w=u=0\); the neural network breaks this algebraic guarantee.

\textbf{3. Uniform coordinate scaling violation (12.2~mHa; tolerance 0.1~mHa).}
The exchange network uses the normalized Laplacian as an input feature:
\begin{codebox}
lapl = _laplacian_from_drho(drho)
kf = fermi_kf(rho, polarized=True)
lapl_norm = lapl / (kf**2 * rho + EPS)
\end{codebox}
Under uniform coordinate scaling \(\mathbf{r}\to\mathbf{r}/\lambda\), the dimensionless descriptors \(s\) and \(t\) are invariant, but \(\mathrm{lapl\_norm}\) is not: the Laplacian of a molecular density does not scale identically to \(k_F^2\rho\) away from the uniform-gas limit. Because the neural network mixes \(\mathrm{lapl\_norm}\) with \((w,u)\), the entire exchange enhancement factor acquires spurious \(\lambda\)-dependence.

\textbf{4. Grid stability violation (max $|\Delta E_{\mathrm{xc}}|=0.100$~kcal/mol; toleran1ce 0.015~kcal/mol).}
Two architectural choices make the functional acutely sensitive to quadrature resolution, as revealed by comparing the (99,590) and (250,974) grids on the AE18 atomization set.
First, the exchange enhancement factor depends on a numerical Laplacian:
\begin{codebox}
lapl = _laplacian_from_drho(drho)   # jnp.gradient on each component
lapl_norm = lapl / (kf**2 * rho + EPS)
\end{codebox}
\texttt{\_laplacian\_from\_drho} applies \texttt{jnp.gradient} (a finite-difference stencil) to each Cartesian component of $\nabla\rho$ on the integration grid. Because the stencil width and point spacing change when the radial or angular grid is refined, \(\mathrm{lapl\_norm}\) is not a true pointwise density functional; it is an inter-grid-point numerical derivative whose value shifts with quadrature resolution. This shift propagates through the sigmoid network into $g_x$ and ultimately into the integrated exchange energy.
Second, the same-spin correlation adds a correction controlled by a steep sigmoid switch in $r_s$:
\begin{codebox}
switch_ss = 1 / (1 + jnp.exp(10 * (rs - 0.5)))
g_aa = g_aa + switch_ss * (p1*w + p2*u + p3*w*u)
\end{codebox}
The slope of 10 produces a peak derivative of $\sim$2.5 near $r_s=0.5$, which falls squarely in the valence-density regime of AE18 atoms. Small shifts in $\rho$ between grids translate into $r_s$ shifts that are amplified by the switch, toggling the correction on or off in slightly different spatial regions. Together, the two mechanisms drive the maximum $|\Delta E_{\mathrm{xc}}|$ to 0.100~kcal/mol, nearly seven times the 0.015~kcal/mol threshold recommended by Refs.~\cite{mardirossian2016wB97MV,liang2026coach}.

\section{Ablation Study: Creativity as a Critical Ingredient}
\label{sec:ablation-creativity}

This report argues that genuine improvement in this search required structural exploration rather than incremental exploitation, and credits multi-island topology and exploration-heavy parent selection as the important mechanisms for maintaining that diversity (Section~\ref{sec:discussion}). Here, we present the outcome of removing these mechanisms, using only a single island and reverting to a 20/80 exploration split. The result was a functional that achieved the \textit{lowest} recorded WRMSD$_{val}$ of any candidate across all runs, yet generalizes catastrophically to the test set. 

\begin{table}[H]
\centering
\small
\begin{tabular}{lccc}
\toprule
\textbf{Functional} & \textbf{WRMSD}$_{\mathrm{val}}$ & \textbf{WRMSD}$_{\mathrm{test}}$ & \textbf{Gap} \\ \midrule
\(\omega\)B97M-V & 4.02 & 3.64 & $-$0.38 \\ 
\texttt{SAFS26-a} & 3.65 & \textbf{3.41} & $-$0.24 \\ 
\texttt{SAFS26-b} & 3.78 & 3.44 & $-$0.34 \\ 
\texttt{WB97MVCrossTerm\ldots ActivatedCSS\_v1} & \textbf{3.64} & 4.09 & \textbf{+0.45} \\ 
\bottomrule
\end{tabular}
\caption{Validation, test, and val--test gap WRMSD in kcal/mol for selected functionals. \texttt{WB97MV\_CrossTerm\_Gated\_TauW\_MultiGate\_RatOSC\_ActivatedCSS\_v1} achieves a record-low validation score while simultaneously producing the widest positive gap of any candidate.}
\label{tab:ablation-gap}
\end{table}

\texttt{WB97MV\_CrossTerm\_Gated\_TauW\_MultiGate\_RatOSC\_ActivatedCSS\_v1} exhibits the most severe validation--test gap of any candidate in the study. The reason is clear, supplied with no outside ideas and repeatedly viewing the same small set of high performing solutions, any large structural change is almost certain to degrade performance. Thus, agents chose `safer' smaller changes. Among other micro-exploitations, the most striking was an emergent strategy of systematically freezing high performing parent's parmeters and grafting a minimal correction on top. Of the 65 total numeric parameters in the functional, \textbf{61 were inherited verbatim from the parent} at their fully trained values, and only the \textbf{4 elements of \texttt{c\_css\_act}} were free to move during L-BFGS training. The structural modification multiplied the existing same-spin enhancement factor by a ratio of two sigmoid activations, a correction which amounts to very little grafted onto an otherwise fixed 61-parameter form. Any improvement over the parent's validation score is thus driven by overfitting to the validation distribution rather than by genuine representational change.

\noindent\textbf{Significance for the creativity claim.}
\texttt{WB97MV\_CrossTerm\_Gated\_TauW\_MultiGate\_RatOSC\_ActivatedCSS\_v1} came from Seed~2.0, the backend that otherwise produced structurally diverse proposals and the two best-generalizing functionals in the study. Its failure therefore cannot be attributed to the intrinsic conservatism of the LLM backend. Rather, it demonstrates that naively iterating on top-performing solutions collapses the entire population diversity in a negative feedback loop. The emergent frozen-parameter strategy is emblematic of this collapse: it offers a near-zero-risk way to guarantee no regression on validation while potentially squeezing out marginal score gains. Without structural mechanisms that force the search away from the current elite even an otherwise high performing SOTA models still converge to this degenerate regime at sufficient search depth.

\begin{tcolorbox}[colback=gray!5,colframe=gray!75!black,boxrule=0.5pt,arc=2mm]
\small
\begin{verbatim}
class WB97MV_CrossTerm_Gated_TauW_MultiGate_RatOSC_ActivatedCSS_v1(
        SeparatedPxExcFunc):
    def __init__(self, trainable_params=("c_css_act",)) -> None:
        ...
        params.update(dict(
        c_x_0 = 0.85,                                              # frozen (UEG constraint)
        c_x_train    = jnp.array([0.46467883,  3.27078968]),       # frozen 
        c_css        = jnp.array([1.10881579, -1.72331564,
                                 -7.26001241, 16.62646155,
                                 -1.15694472]),                     # frozen 
        c_cos        = jnp.array([0.24988053, -9.39610915,
                                  6.10997423, -9.37973002,
                                -15.28420953, 16.03908365]),        # frozen 
        c_x_cross    = jnp.array([-0.12446728, -2.87756573]),      # frozen 
        c_css_cross  = jnp.array([ 0.98494613]),                   # frozen 
        c_cos_cross  = jnp.array([ 1.40793417,-12.8656106 ]),      # frozen 
        c_x_gate     = jnp.array([-0.03675481,  0.50439803]),      # frozen 
        c_css_gate   = jnp.array([-2.29262665]),                   # frozen 
        c_cos_gate   = jnp.array([14.47816245, 26.0946234,
                                   0.45369871]),                    # frozen 
        c_x_tw       = jnp.array([-0.26107215, -0.12455014]),      # frozen 
        c_css_tw     = jnp.array([14.58461232]),                   # frozen 
        c_cos_tw     = jnp.array([-2.56354756,  1.05986868,
                                  -0.939609  ]),                    # frozen 
        c_x_gate_tw  = jnp.array([ 0.35455838,  2.1550685 ]),     # frozen 
        c_css_gate_tw= jnp.array([-0.25252967]),                   # frozen 
        c_cos_gate_tw= jnp.array([ 0.37069902]),                   # frozen 
        c_cos_tw_coupled = jnp.array([1.97558549,  0.24267317]),   # frozen 
        c_rat_num    = jnp.array([-0.24245632,  0.07568424,        # frozen 
                                  -0.58029248,  0.0038596 ,
                                  -0.13676558,  0.05632591,
                                   0.02322058, -0.02319411]),
        c_rat_den    = jnp.array([-0.24487811, -0.15191397,        # frozen 
                                  -0.54878029, -0.00818291,
                                  -0.01495359,  0.01088918]),
        c_rat_num_ext2 = jnp.array([ 0.00163673, -0.00321903]),    # frozen 
        c_rat_den_ext2 = jnp.array([-6.44e-04,    1.74e-05 ]),     # frozen 
        c_osc_elf    = jnp.array([ 0.82254053,  1.40289355,        # frozen 
                                  -0.20961353]),
        c_osc_kingrad= jnp.array([ 0.0,  0.0,  0.0]),             # frozen 
        # ---- only trainable parameters ----
        c_css_act    = jnp.array([ 0.0,  0.0,  0.0,  0.0]),       # TRAINABLE, zero-initialized
        ))
        ...
\end{verbatim}
\end{tcolorbox}

\end{document}